\documentclass[letterpaper, 10 pt, journal, twoside]{IEEEtran}

\IEEEoverridecommandlockouts

\usepackage{graphicx}
\usepackage{epsfig}
\usepackage{mathptmx}
\usepackage{times}
\usepackage{amsmath}
\usepackage{amssymb}
\usepackage{hyperref}
\usepackage{cite}

\title{Motion optimization and parameter identification for a human and lower-back exoskeleton model}

\author{Paul Manns$^{\dagger}$, Manish Sreenivasa$^{\dagger}$, Matthew Millard$^{\dagger}$ and Katja Mombaur$^{\dagger}$%
\thanks{Manuscript received: September, 10, 2016; Revised December,
21, 2016; Accepted February, 7, 2017.}%
\thanks{This paper was recommended for publication by Ken Masamune 
upon evaluation of the Associate Editor and
Reviewers' comments. Financial support by the European Commission within the H2020 project SPEXOR (GA 687662) is gratefully acknowledged.}%
\thanks{$^{\dagger}$The authors are with the Interdisciplinary Center for Scientific Computing, Heidelberg University, 69120 Heidelberg, Germany 
{\tt\small \{paul.manns, manish.sreenivasa, matthew.millard, katja.mombaur\}@iwr.uni-heidelberg.de}}%
\thanks{
This is the author's version of an article that has been published in this journal. Changes were made to this version by the publisher prior to publication.
The final version of record is available at: \url{http://dx.doi.org/10.1109/LRA.2017.2676355}}}

\begin{document}
\maketitle
\markboth{IEEE Robotics and Automation Letters. Preprint
Version. Accepted February, 2017}{Manns \MakeLowercase{\textit{et al.}}:
Motion optimization for a human and exoskeleton model}
\begin{abstract}
Designing an exoskeleton to reduce the risk of low-back injury during lifting is challenging.
Computational models of the human-robot system coupled with predictive movement simulations can help to simplify this design process. 
Here, we present a study that models the interaction between a human model 
actuated by muscles and a lower-back exoskeleton. 
We provide a computational framework for identifying the spring parameters of the exoskeleton using an optimal control approach and forward-dynamics simulations. 
This is applied to generate dynamically consistent bending and lifting movements in the sagittal plane. Our computations are able to predict motions and forces of the human and exoskeleton that are within the torque limits of a subject. 
The identified exoskeleton could also yield a considerable reduction of the peak lower-back torques as well as the cumulative lower-back load during the movements. 
This work is relevant to the research communities working on human-robot interaction, and can be used as a basis for a better human-centered design process. 
\end{abstract}
\begin{IEEEkeywords}
Prosthetics and Exoskeletons, Optimization and Optimal Control, Physically Assistive Devices
\end{IEEEkeywords}
%
%
\section{Introduction}
\IEEEPARstart{L}{ower}-back pain among workers in physically demanding jobs (e.g. nurses, airport baggage lifters, construction workers) is a frequent cause for absenteeism leading to production losses as well as burdens on the health infrastructure \cite{Hagen1998,Maniadakis2000}. 
In many of these work scenarios the load on the lower back, and the associated risk of injury, can be reduced by wearable robotic devices (exoskeletons) that support, augment or limit motion. 
Several challenges have to be solved before such devices become commonplace. 
\begin{figure}[t]
  \centering 
  \includegraphics[width=0.48\textwidth]{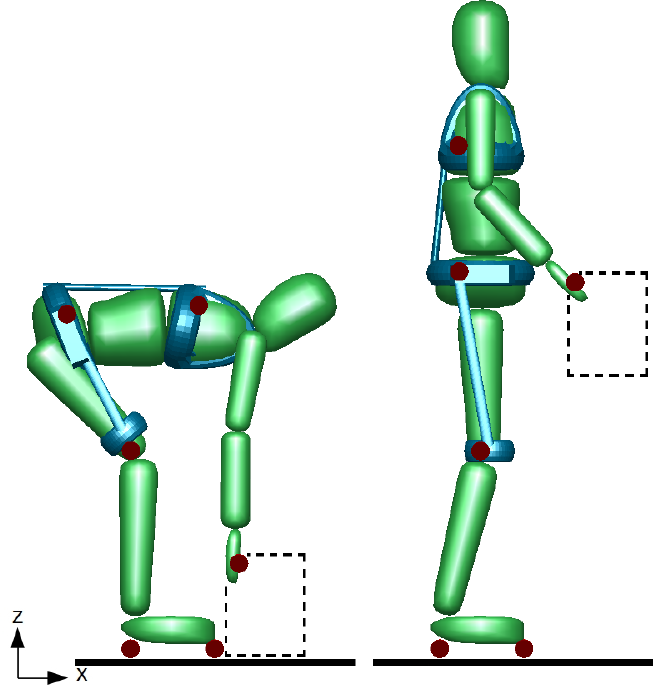}
  \caption{Visualization of a human and exoskeleton model pose during a bending and lifting task. 
  The attachment points of the exoskeleton on the thigh, pelvis and uppertrunk segments of the human body are marked with red circles. 
  The red circles on the hands and feet indicate the contact points used for the lift, and the foot-ground contact points, respectively.} 
  \label{fig:exolifter2d}
\end{figure}

A majority of exoskeleton research and development focuses on the lower-limbs (e.g. \cite{Zoss2006,Talaty2013,Kolakowsky2013,Moreno2014,Beil2015,Wall2015}) 
or the shoulder-arm complex \cite{Stadler2014,PonsBook}. 
Relatively few devices deal with movements of the back and torso. 
The movements and forces required to support the back require a specialized exoskeleton design. 
Some examples of exoskeletons developed for back support are Robomate 
\cite{Stadler2014}, Laevo (Laevo B.V., Netherlands), PLAD \cite{Abdoli2006} and WSAD \cite{Luo2013}. 
The design of the exoskeleton may vary considerably, but in general, the idea is to support the extension moments at the back during bending thereby reducing the muscle forces as well as the corresponding vertebral forces. 
There is a gap in the experimental and simulation literature for studies that examine the motions and forces of supported human lifting. 
Additionally, there is limited knowledge on how we can predict lifting movements and the stresses this applies on the human body.

Here, we present a computational framework to help the design process and customization of lower-back exoskeletons to both the user and the type of movements they are intended to support. 
We chose a stoop motion (Fig. \ref{fig:exolifter2d}) for our case study because it is commonly used when picking something off the ground \cite{Vandieen1999}. 
Back injury is correlated with the integral of net back moments over time 
\cite{Coenen2013}, a quantity known as cumulative lower back load (CLBL), which we are able to compute with the presented framework. 
The study by Dieen et al \cite{Vandieen2005} also found significant correlations between spinal compressive forces and the net moment about the L5/S1 vertebrae. 
Lower-back exoskeletons, such as the one modeled in our present study, aim to reduce the torque required during lifting motions, thereby reducing the risk of high compression forces and back injury.

Our work is in the context of a larger effort within the SPEXOR\footnote{\url{www.spexor.eu}} project, aimed at developing a lower-back 
exoskeleton for vocational reintegration. 
The design of the exoskeleton in this project (as is likely the case for exoskeletons in general \cite{PonsBook}) faces multiple challenges. 
The initial design parameters are only vaguely known and it can take many costly iterations before a design may be finalized. 
Furthermore, the vast range of human body shapes and potential movements makes it very labor-intensive to test all combinations experimentally.

Computational models of the human body and the exoskeleton can help with these issues and complement experimental testing. 
An important prerequisite is to \emph{predict} the natural motion 
of the human body, while carrying our daily tasks such as lifting and walking. 
In this context our current study makes the following contributions:
\begin{itemize}
	\item 
	We develop a framework to model the effects of a passive lower-back exoskeleton worn by a human
	\item We identify the optimal exoskeleton parameters suited for a bending and lifting motion while carrying heavy loads (an object of $15\,\text{kg}$ weight in our scenario)
\end{itemize}
For this initial work, we restrict ourselves to two-dimensional movements in the sagittal plane and a passive exoskeleton mechanism (i.e. only consisting of unactuated mechanical elements such as springs and dampers). 
It is important to note that our framework does not require any recorded 
motion data from the human subjects as input.
Instead, the movement trajectories and the optimal exoskeleton parameters are solutions of an optimal control problem (OCP). 
OCPs, described further in Section \ref{sec:ocp}, allow to predict feasible motions for given dynamic models and suitable objective 
functions. 
They have been used successfully for robot and human motion 
generation in the past (e.g. \cite{Bobrow1985, Vonstryk1994, Mombaur2010}), and to a limited extent for the design of lower limb exoskeletons \cite{Koch2015,Mombaur2016}.
\section{Human and exoskeleton modeling}
We model the human body in the sagittal plane as an articulated mechanism (Fig. \ref{fig:exolifter2d}). 
The dynamics governing this model may be described as:
\begin{align}\label{eq:DAE}
	M(\mathbf{q}) \mathbf{\ddot{q}} + c(\mathbf{q},\mathbf{\dot{q}}) 
	  &= \mathbf{\tau} + G(\mathbf{q})^T \mathbf{\lambda} \\
	\label{eq:DAE2}
	g(\mathbf{q}) &= 0~,
\end{align}
where $\mathbf{q},\mathbf{\dot{q}}$ and $\mathbf{\ddot{q}}$ denote the generalized coordinates, their velocities and accelerations, $M$ the generalized inertia matrix, $c$ the Coriolis, gravitational and centrifugal forces and $\mathbf{\tau}$ the impressed forces. 
$g$ denotes the scleronomic constraint function, $G$ its Jacobian and $\mathbf{\lambda}$ the Lagrange multipliers. 
In our case, the generalized force vector consists of the summation of the agonist-antagonist forces exerted by muscle torque generators (MTGs), $\mathbf{\tau}_M$, as well as the forces exerted by the exoskeleton $\mathbf{\tau}_E$. 
The MTGs and the exoskeleton are described in the following sections. The forward dynamics is evaluated using Featherstone's articulated body algorithm (ABA) and spatial vector algebra as described in 
\cite{Featherstone1983,Featherstone2014}. 
We employ the open source dynamics library \textit{Rigid Body Dynamics Library (RBDL)} \cite{Felis2016}, which implements ABA.
\subsection{Human body model}
Our human model is a kinematic tree consisting of $16$ segments with a rotational joint between each segment. 
The pelvis is modeled with two additional translational degrees of freedom (DoFs) in the forward (x) and vertical (z) directions (Fig. \ref{fig:exolifter2d}). 
We use the anthropometric parameters of a $35$ year old male subject weighing $77.5\,\text{kg}$ with a height of $1.72\,\text{m}$. 
The segment lengths, segment masses and inertia properties were computed based on the linear regression equations from De Leva \cite{Deleva1996}. 

Except for the DoFs at the pelvis, all the other $15$ rotational joints of 
the model are actuated by pairs of agonist-antagonist MTGs, giving $30$ actuators in total. 
The MTGs assume a rigid tendon and represent the overall torque 
being generated at a joint by the muscles in one rotational direction. 
In comparison to line-type muscles (see e.g. \cite{Delp2007:OpenSim}), this simplification significantly reduces the computational time while preserving the experimentally measured torque-angle and torque-angular-velocity characteristics \cite{Anderson2007,Jackson2010,Kentel2011}. 
Our MTG model computes the joint torque as a function of joint angle, 
angular velocity, and muscle activation as
\begin{align}\label{eq:mtg}
	\tau_M = \tau_0 \left( a	f^A(q) f^V(\dot{q})  
	 - \beta \frac{\omega}{\omega_{\text{max}}} 
     + f^{P}(q)  
	 \right)
\end{align}
for either agonist or antagonist of a certain joint, where $\tau_0$ indicates the 
maximum isometric torque, $a$ denotes the normalized muscle activation ranging 
from 0 (no activation) to 1 (full activation), $f^A$ denotes the active 
torque-angle curve, $f^V$ denotes the active-torque-velocity curve and $f^P$ 
the passive-torque-angle curve. 
Damping is specified with the $\beta \omega / \omega_{\text{max}}$ term. 
We have fitted the curves for $f^A$, $f^V$, $f^P$ as $5^{th}$ order Bezier 
splines. 
They are built on dynamometric data from the literature, primarily 
relying upon \cite{Anderson2007,Jackson2010,Kentel2011}.
These curves are twice continuously differentiable which allows their 
use inside the OCP. 
The MTG implementation is available as open source software code\footnote{\url{https://bitbucket.org/rbdl/rbdl}}, with further details in \cite{Millard2017}.
\begin{figure}[thpb]
  \centering 
  \includegraphics[width=0.48\textwidth]{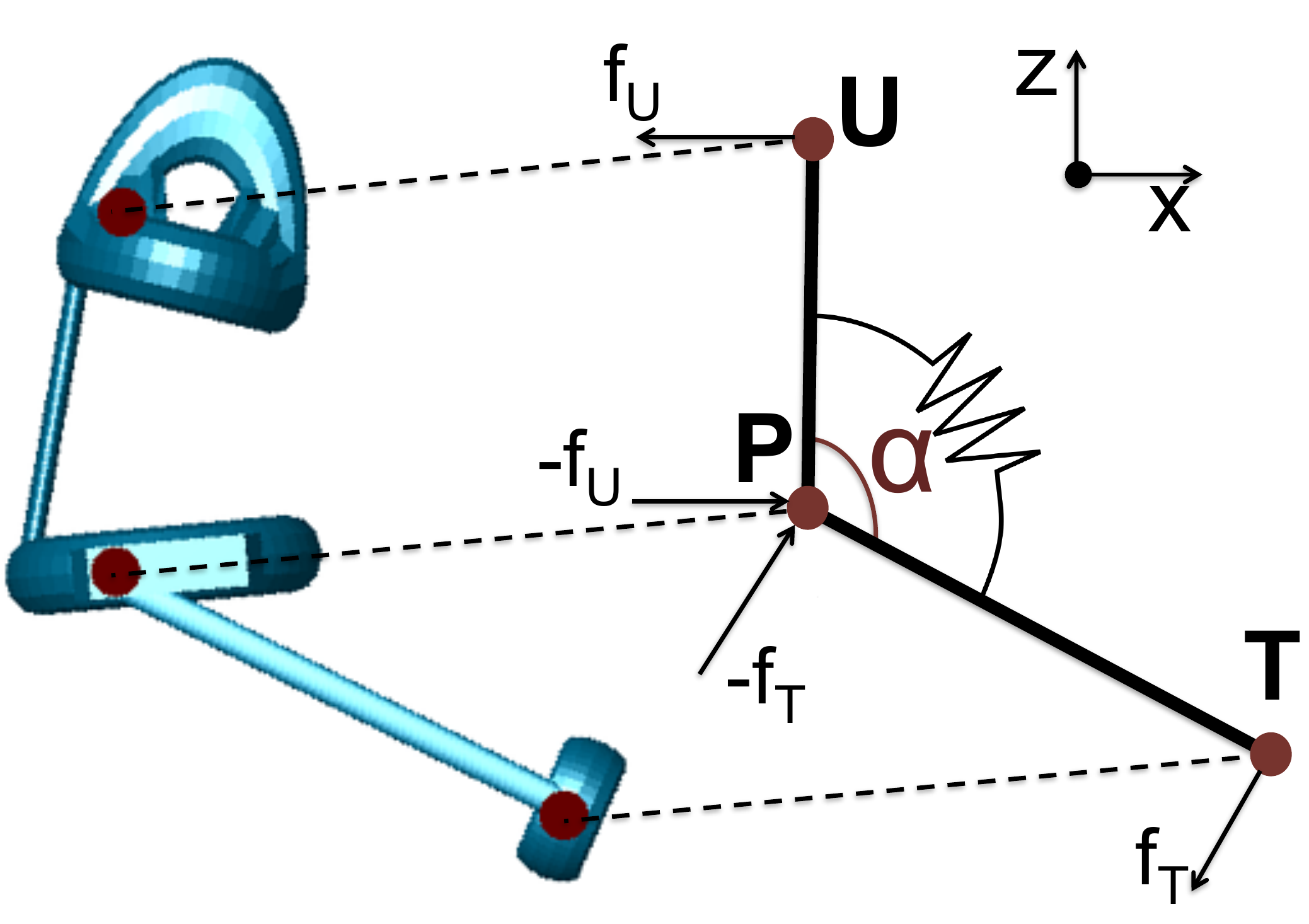}
  \caption{Sketch of the exoskeleton mechanism in the sagittal plane. Forces and 
  the directions they are acting in are marked with arrows. The attachment points 
  are marked with filled red circles.} 
  \label{fig:exomech}
\end{figure}
\subsection{Exoskeleton model}\label{sec:exo}
We model the exoskeleton as a torsion spring that applies a moment at the point P 
on the pelvis and attaches to the body and thigh at the points T and U, 
respectively (Fig. \ref{fig:exolifter2d}, \ref{fig:exomech}). 
We assume that the exo is securely fastened to the body and does not slip. 
The rigid bars that span the points P-T and P-U are fixed at point P, but the ends are free to slide at the points T and U. 
The torsion angle $\theta$ of the spring is given by
\begin{align}
	\theta &= (\pi - \alpha) - \theta_0 \
\end{align}
where $\alpha$ is the angle extended between the points T, P and U. 
$\theta_0$ denotes the angle where the torsion spring is at rest. 
Considering an ideal spring, the resulting torque is
\begin{align}
	\tau_{\text{spring}} &= k \theta
\end{align}
where $k$ denotes the stiffness of the spring. The resulting forces acting on the attachment points are given by $r \times f = \tau$. As the attachment points lie in the sagittal plane and the force acts perpendicular to the sliding bar on the attachment point, the forces acting on the attachment points can be computed as:
\begin{align}
	f_{U} &= \frac{\tau_{\text{spring}}}{\|\vec{PU}\|}\ 
	         \frac{\vec{PU} \times e_y}{\|\vec{PU} \times e_y\|} \\
	f_{T} &= -\frac{\tau_{\text{spring}}}{\|\vec{PT} \|}\ 
	          \frac{\vec{PT} \times e_y}{\|\vec{PT} \times e_y\|} 	
\end{align}
The Jacobian of the position of point A on the human model with respect to q, the vector of generalized positions, is denoted $J(q,A)$. 
The generalized forces that arise from the exoskeleton mechanism are:
\begin{align}
	\mathbf{\tau}_E &= J(\mathbf{q}, P) (-f_T - f_U) 
	        + J(\mathbf{q}, T) f_T
	        + J(\mathbf{q}, U) f_U
\end{align}
The computation for the spring mechanism on the other side of the body is identical. The resulting generalized torque $\mathbf{\tau}_E$ is added to the torque vector arising from the MTGs before the sum is applied to the rigid body model. 

The angle $\theta_0$ and the spring stiffness $k$ make up the two design parameters 
that are to be identified in the OCP. We set angular limits such that 
physiologically unrealistic extension of the body is not possible. This is the 
same as assuming that the exoskeleton is physically limited by hardstops in 
extension. This is necessary because making $\theta_0$ subject to the optimization 
can yield a preloaded spring generating an extension moment when standing straight.

We assume an exoskeleton mass of $6.6\,\text{kg}$ similar to existing passive 
exoskeletons \cite{Ikeuchi2009}. We distribute this mass to the pelvis, left thigh, 
right thigh, middletrunk and uppertrunk segments with the ratios 
$1/2, 1/8, 1/8, 1/8$ and $1/8$ respectively. 
Note that in this first approach we add the exoskeleton mass to the body segments, and ignore some of the mechanical effects and restrictions the exoskeleton imposes on the motion.
\begin{figure}[thpb]
  \centering 
  \includegraphics[width=0.48\textwidth]{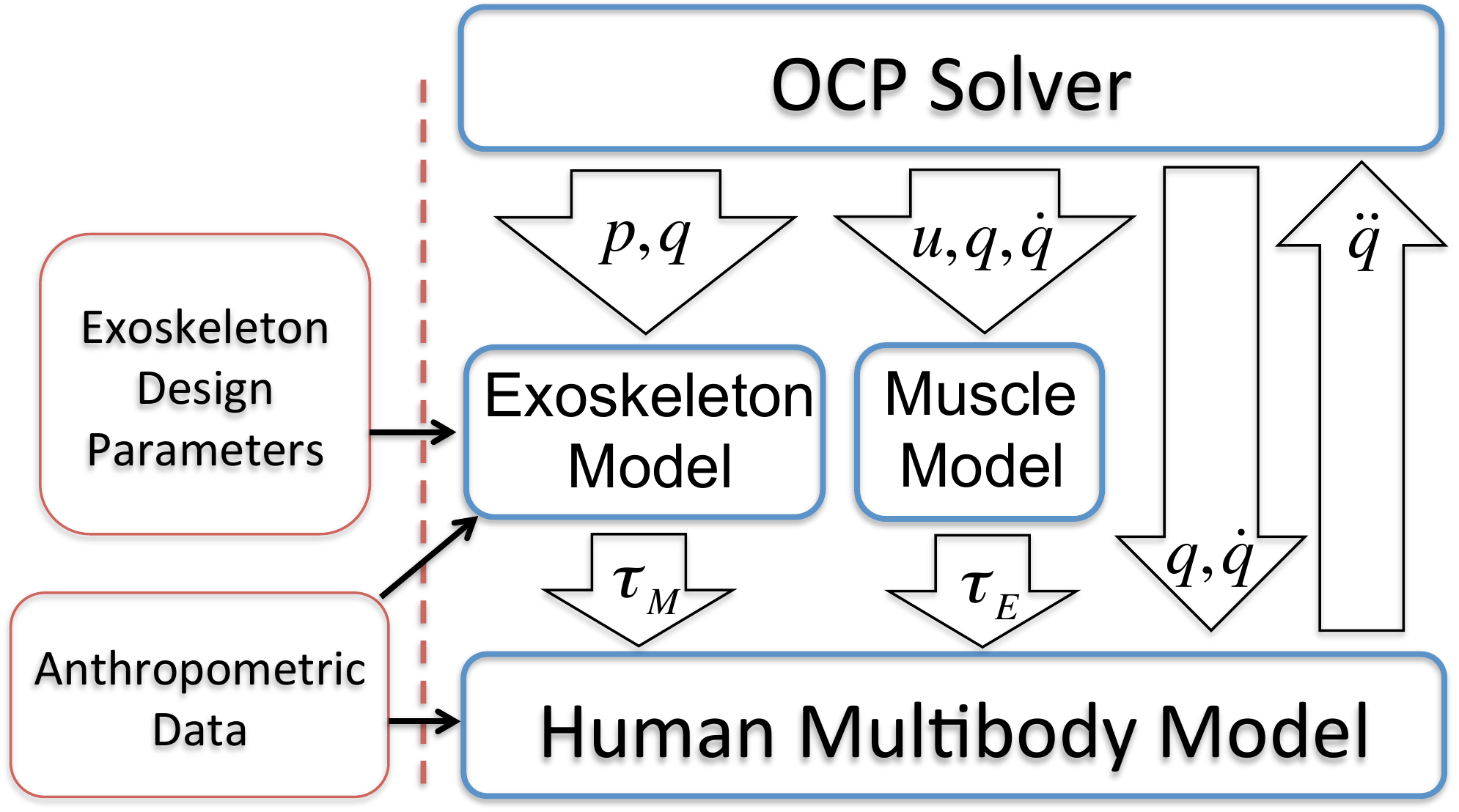}
  \caption{Overview of the OCP setup: The differential state vector of the OCP consists of the generalized 
  coordinates $\mathbf{q}$ and the corresponding velocities $\mathbf{\dot{q}}$. 
  The controls are denoted $\mathbf{u}$ and the global parameters are denoted 
  $\mathbf{p}$. $\mathbf{\tau}_M$ are the generalized forces from the muscle 
  torque generators and $\mathbf{\tau}_E$ the generalized forces arising from 
  the exoskeleton model.} 
  \label{fig:overview}
\end{figure}
\section{Motion and parameter optimization}\label{sec:ocp}
We formulate our OCP (overview in Fig. \ref{fig:overview}) as a $N$ phase problem of the form:
\begin{align}\label{eq:OCP}
	\min_{\mathbf{x},\mathbf{u},\mathbf{p},\mathbf{T}} 
	\sum_{i = 0}^{N - 1} 
	\int_{T_i}^{T_{i + 1}} \phi_i (\mathbf{x}(t), \mathbf{u}(t), \mathbf{p}) 
	~{\rm d}t
\end{align}
s.t.
\begin{align}
\mathbf{\dot{x}} &= f_i(\mathbf{x}(t), \mathbf{u}(t), \mathbf{p})
    ~,~t \in [T_i, T_{i+1}] \label{eq:dynamics}\\
0 &\leq g_i(\mathbf{x}(t), \mathbf{u}(t), \mathbf{p})
    ~,~t \in [T_i, T_{i + 1}] \\
0 &= h_i(\mathbf{x}(t), \mathbf{u}(t), \mathbf{p})~,~t \in [T_i, T_{i + 1}] \\
0 &\leq r_{ieq}(\mathbf{x}(T_0), \ldots, \mathbf{x}(T_{N - 1}), \mathbf{p}) \\
0 &= r_{eq}(\mathbf{x}(T_0), \ldots, \mathbf{x}(T_{N - 1}), \mathbf{p}) \\
&~~ 0 = T_0 < \ldots < T_{N - 1} = T_f
\end{align}
The differential states of the system are denoted $\mathbf{x}$, the controls 
are denoted $\mathbf{u}$, and the global parameters are denoted $\mathbf{p}$. 
The global parameters $\mathbf{p}$ allow us to incorporate the exoskeleton design 
parameters $\theta_0$ and $k$. 
The variables $T_i~,~i \in \{0, \ldots, N-1\}$ are the start and end times of 
the different phases and also subject to the optimization. 
Equation \eqref{eq:OCP} is the minimization objective $\phi_i$. The function $f_i$ in Eq. \eqref{eq:dynamics} describes the system dynamics.  Our human model has a floating base and consequently, we impose equality constraints $h_i$ modeling the heel and toe positions which are kept in contact with the floor throughout the motion. In addition, we impose inequality constraints $g_i$ to maintain positive contact forces at these points. Constraints at the start, lift and end points such as the hand positions in task space enter the boundary constraints $r_{ieq}$ and $r_{eq}$. Note that our system is scleronomous and hence, $f_i$, $g_i$ and $h_i$ do not depend on $t$ explicitly. 

The problem described above contains significant nonlinearities and is hard to 
solve. The nonlinearities stem from the computation of the generalized inertia 
matrix, the MTG computations and the constraints taking care of avoiding 
interpenetrations. 
In particular, the set of feasible solutions is non-convex in general and we can 
only expect to obtain a local solution when computing the solution numerically. 
We use a direct solution approach that discretizes the controls in the time 
domain as piecewise linear functions. Direct multiple shooting \cite{Bock1984} is 
a popular approach for solving such problems. It is based on a subdivision of the 
phases into shooting intervals on which initial value problems of the 
ordinary differential equations are solved depending on the current state, 
control and parameter iterate. We employ the SQP-based direct multiple shooting 
implementation MUSCOD-II \cite{Leineweber2003} to solve the OCPs described here.
\subsection{Objective function}
For simulating the motions in this work we use an objective function that 
minimizes the squared muscle activations over time. 
This formulation is commonly used in literature on optimal control for human motion generation and is associated with the minimization of muscle effort \cite{Ackermann2010}. 
In our case the muscle activations $\mathbf{a}$ are the controls $\mathbf{u}$ of the OCP, hence we may formulate our objective function as:
\begin{align}\label{eq:obj}
	\min 
	\sum_{i = 0}^{N - 1} 
	\int_{T_i}^{T_{i + 1}} \mathbf{u}(t) \cdot \mathbf{u}(t)
	~{\rm d}t
\end{align}
\subsection{Movement scenarios}
The movement we consider for the trajectory optimization consists of $N = 2$ phases, namely bending and lifting. 
The human model starts from an upright pose while standing still (all velocities to zero). 
At the start of the second phase the model's hands are constrained to be at a lifting point directly in front and close to the ground. 
At the end of the second phase the model is required to be in an upright pose while standing still. 
We introduce a slight left-right asymmetry in the initial pose and the lift point as in real human movements perfect symmetry is unlikely. 
This allows the optimization routine to find non-symmetric solutions which would be not possible otherwise.
We compute four movements: 
$0\,\text{kg}$ and $15\,\text{kg}$ lift both with and without exoskeleton.
The $0\,\text{kg}$ lift was chosen because many agricultural workers pick fresh crops and suffer from back pain \cite{Fathallah2010}. 
The $15\,\text{kg}$ lift was chosen because it is in the middle of the \emph{heavy} category for frequent lifting \cite{Mooney1996}.
Importantly, this category of workers experience significantly more back pain than workers in the medium category.
\subsection{Multiple Shooting Setup}
We discretize both phases with $25$ shooting nodes and $25$ control intervals. 
Both phases are initialized with a duration of $1\,\text{s}$. The start and end 
configuration of the model for both phases are initialized with predefined feasible 
poses in upright and bent position. 
The intermediate trajectories are then initialized as a linear interpolation 
between the given poses. All joint velocities are initialized as $0$. 
For all control trajectories, we have chosen piecewise linear, globally continuous 
splines. 
They are initialized with a muscle activation of $0.25$ for the whole time horizon.
\begin{table}[htpb]
\caption{Switching and final time of the optimized motion for $0\,\text{kg}$ and $15\,\text{kg}$ lifts with and without exoskeleton assist}
\label{tbl:times}
\begin{center}
\begin{tabular}{|r||c|c|c|c|}
\hline
 & $0\,\text{kg}$ no exo & $0\,\text{kg}$ exo
 & $15\,\text{kg}$ no exo & $15\,\text{kg}$ exo \\
\hline
\hline
$T_{\text{switch}}$ & 0.80 & 0.98 & 0.99 & 0.80 \\
$T_{\text{f}}$ & 2.00 & 2.18 & 2.15 & 1.79  \\
\hline
\end{tabular}
\end{center}
\end{table}
\begin{figure*}[thbp]
  \centering 
  \includegraphics[width=0.95\textwidth]{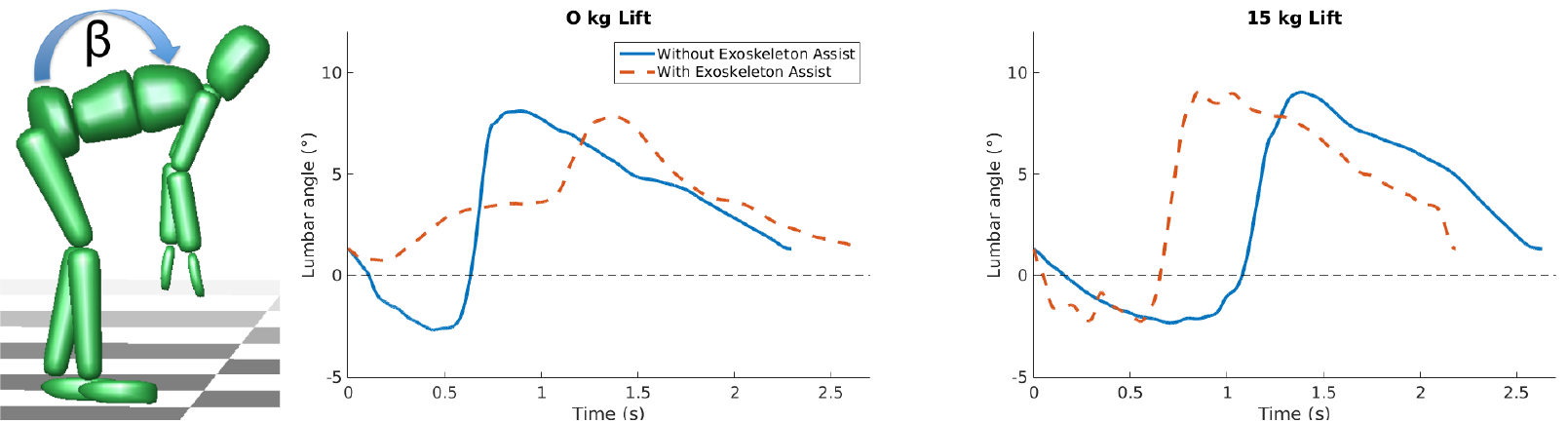}
  \caption{Progression of lumbar angle, $\beta$, during the movements with and without exoskeleton assist for $0\,\text{kg}$, and $15\,\text{kg}$ lifts.} 
  \label{fig:lumbarAngles}
\end{figure*}
\section{Results}
Each of the four OCPs took between 4 to 8 hours to solve on a standard desktop computer (Intel(R) Core(TM) i7-4790K CPU with $4.00$\,GHz running with $4$ compute threads).
\subsection{Movement Profiles}
For all cases the predicted movements showed smooth profiles and stayed within the limits defined for joint angles, joint angular velocities as well as the generated muscle torques. 
We observed that the lumbar angle $\beta$, the angle between the 
pelvis and the upper trunk, showed greater variation with the exoskeleton for the 15kg lift, than while lifting no weight (Fig. \ref{fig:lumbarAngles}). 
The maximum angular difference between the lumbar angles was $11.2^\circ$, the one at the switching point was $0.81^\circ$, the difference between peak lumbar angles was $1.25^\circ$. 
The durations of the movements differed depending on the lifting weight as well as whether the exoskeleton assist was available or not (see Table \ref{tbl:times}). The largest time difference was found for the 15kg lift, which lasted 2.15s without the exoskeleton and 1.79s with the exoskeleton.
\subsection{Lumbar extension torque and cumulative load}
\begin{figure}[thpb]
  \centering 
  \includegraphics[width=0.4\textwidth]{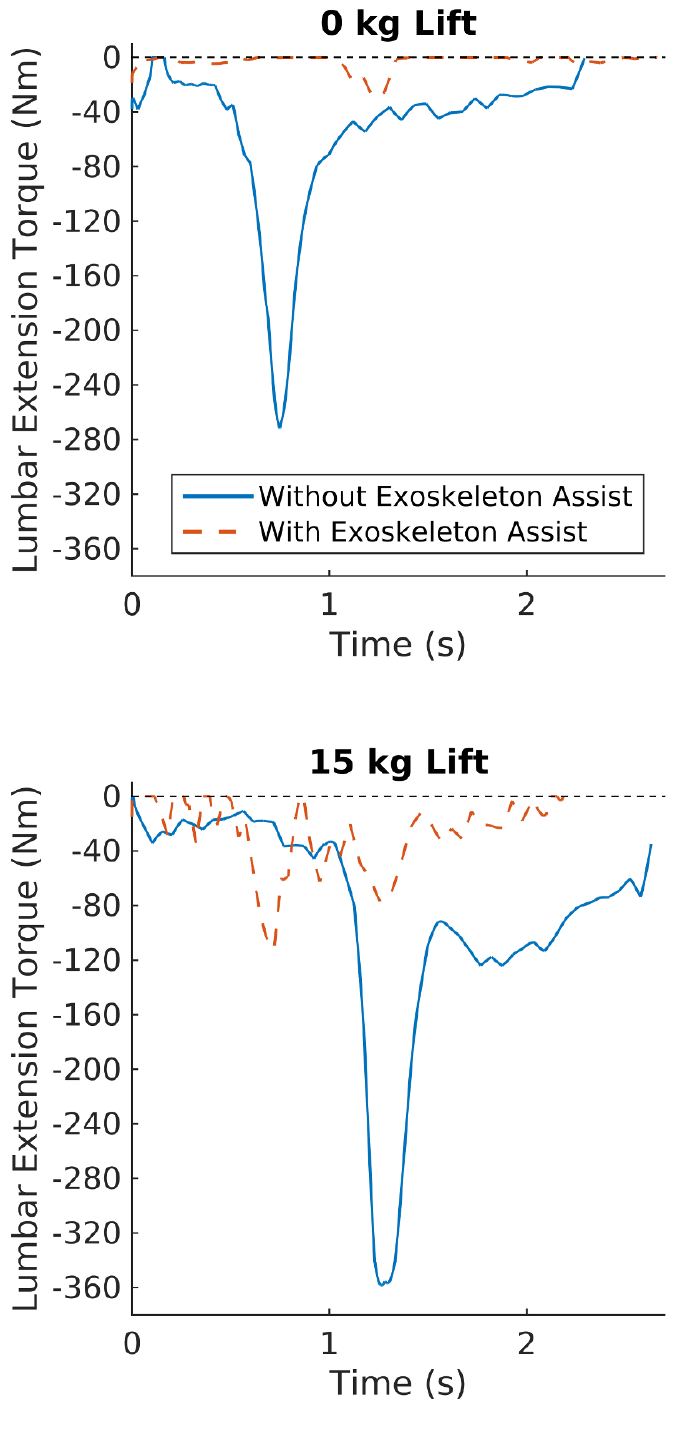}
  \caption{Profile of the muscle extension torques at the lower back with and without exoskeleton assist for (a) $0\,\text{kg}$ and 
           (b) $15\,\text{kg}$ lifts.} 
  \label{fig:lumbarTorques}
\end{figure}

The exoskeleton assist considerably reduces the lower-back extension torques exerted by the model (Fig. \ref{fig:lumbarTorques}). 
The peak torque was lowered from $271.6\,\text{Nm}$ to $29.5\,\text{Nm}$ in the $0\,\text{kg}$ case, and from $358.3\,\text{Nm}$ to $112.3\,\text{Nm}$ in the $15\,\text{kg}$ case. 
The associated muscle effort (as measured by the objective function value) required for the lift also reduced significantly when the model was wearing an exoskeleton; by $76.4\,\%$ in the $0\,\text{kg}$ case and by $51.9\,\%$ in the $15\,\text{kg}$ case. CLBL reduced by $93.8\,\%$ from $117.54\,\text{Nm\,s}$ to $7.29\,\text{Nm\,s}$ for the $0\,\text{kg}$, and by $79.28\,\%$ from $222.14\,\text{Nm\,s}$ to $46.02\,\text{Nm\,s}$, for the $15\,\text{kg}$ case.
\subsection{Optimal exoskeleton parameters}
The optimal spring offset angle, $\theta_0$, were found to be $-8.88^\circ$ and $-9.28^\circ$ for the $0\,\text{kg}$ and $15\,\text{kg}$ lift, respectively. Note that for our setup, $\theta_0 = 5.9^\circ$ would yield a completely unloaded spring when standing upright. The OCP results therefore favor preloading the spring to generate small extension moments when the human is standing upright. The optimized spring stiffness for the $0\,\text{kg}$ lift was found to be $72.1\,\text{Nm}/\text{rad}$ and that for the $15\,\text{kg}$ lift was $31\,\%$ lower at $50.0\,\text{Nm}/\text{rad}$.  
\section{Discussion}
In this study we have presented a framework to model the interaction between a 
human and a lower-back exoskeleton. 
Additionally, we predict the movements during 
a bending and lifting task using an optimal control approach and identify the 
exoskeleton spring parameters for these movements. 
Our results provide the first steps towards applying model-based predictive simulations of human-robot interaction to design wearable robotic devices. 
To the best of our knowledge such approaches are not yet widely used with few comparable studies (see e.g. \cite{Koch2015,Mouzo2016,Wang2011,Mombaur2016}). 
We note the following 
important points about our present work:
\begin{itemize}
\item
The motion computation requires minimal input about the human operator (height and weight), pre-defined motions (kinematic pose at start, lift and end), as well as the exoskeleton design. 
This aspect can help to reduce the time and effort required in the initial design phases.
\item
The OCP framework produces (\emph{predicts}) dynamically consistent motions for the whole simulation. 
With the use of MTGs for the human, and feasible mechanical models of the exoskeleton, the results are also physiologically meaningful. 
\end{itemize}

To transfer our approach to the exoskeleton design process; we may imagine that a large number of such simulations would be carried out on a variety of tasks, lifting weights, human body proportions as well as exoskeleton designs. 
The result of this process may then be one or several exoskeleton designs 
that are best able to satisfy the design requirements (e.g. reduce lower-back torques by $40\,\%$).

Our results show that the OCP is able to identify exoskeleton parameters that reduced both the peak torques at the lower-back and CLBL. 
This reduction is desirable as high peak lower-back torques and CLBL is associated with lower-back pain and injury \cite{Coenen2013,Vandieen2005}. 
Note that both of these quantities were not tied to the chosen objective function. Comparing to literature, our simulations yielded peak back extension torques that were higher than those reported in \cite{Kingma2004} ($221 \pm 24\text{Nm}$ for a $10\,\text{kg}$ lift). As well, the flexion angle reported in \cite{Kingma2004} ($39 \pm 15^\circ$) is higher than the ones in our model ($8-9^\circ$). We expect that appropriate changes to the cost function could align our simulation results more closely with those from \cite{Kingma2004}. 
The reason for these differences is likely because we treated muscle activation as being equally expensive across all muscles.
If the activation of the back muscles were to be weighted as more expensive, relative to other muscles, the lumbar spine would go into deeper flexion. As the variation in lumbar flexion angle can be quite variable ($\sigma=15^\circ$ \cite{Kingma2004}) the weighting pattern likely varies from one subject to the next.
The peak lumbar torque can be reduced by including the total torque the muscle generates in the cost function. Under an activation squared cost function the passive forces of the hip extensors (which the solution exploits) are without cost. 
Inverse optimal control can provide means to identify the cost function and weighting terms from experimental data \cite{Mombaur2016}, and is one of avenues we are currently exploring.

The optimal parameters of the exoskeleton varied with the weight being lifted as did the value of the cost function. 
As expected the exoskeleton assisted motion resulted in a large reduction in the cost function value. 
Interestingly, the heavier weight yielded a less stiff spring while the rest lengths were similar for both 0kg and 15kg lifts. 
The 15kg lift requires a lighter spring because the passive component of the hip extensors are being exploited to carry more of the load. 
This observation has important ramifications: the ideal spring coefficent for a subject is likely influenced by the flexbility of their hips. 
Since flexbility varies greatly from one person to the next, it is important that the stiffness characteristics of the exoskeleton be easily adjusted.
Interestingly, the time of the motion with and without exoskeleton
showed different trends for 0kg and 15kg weight. 
The reason for these differences are presently unclear and may be related to the cost function formulation as well as OCP solver exploiting the cost-free passive muscle forces. 
These issues are the subject of further investigation, as well as comparisons to experimental lifting data. 
\addtolength{\textheight}{-1.5cm} 
\subsection*{Limitations and Perspectives}
This work is an initial approach towards model-based design of spinal exoskeletons. To focus on the development of the framework we have introduced several model simplifications that must be taken into consideration. Here we list some of these limitations, their possible influence on our results, and perspectives towards improving our current framework in this context.

The objective function used in this study is typically used for the generation of gait \cite{Anderson2007}. The extent to which this is applicable to lifting motions is presently unclear. Additionally, neither our current cost function, nor the constraints, account for ergonomic issues such as comfort and pain (e.g. from pressure at contact points). Incorporating an ergonomically oriented component in the OCP is an important next step towards designing exoskeletons that will be acceptable by the end-users. 

Personalization of the human and exoskeleton model to the user is another vital requirement. Here we developed a human model (and the corresponding  exoskeleton) scaled to the proportions of our single test subject. In current work we are developing methods to identify the body parameters from experimental data, in order to better match the body proportions and physical strengths of a range of subjects.

The movements limited to the sagittal plane and our simplified exoskeleton model provide only a gross approximation of the real-world lifting scenarios. While our current work illustrates the potential of optimal control methods for exoskeleton design, further developments are necessary to make them usuable in a general sense. For example, extending the human and exoskeleton models to movements in the transverse and coronal planes, and, modeling the exoskeleton as an external kinematic chain coupled to the body at the contact points. Finally, these perspectives should be complemented with validation of the methodology including experimental trials with a cohort of healthy subjects lifting with and without an exoskeleton.
\section*{ACKNOWLEDGMENT}
The authors would like to thank the Simulation and Optimization
research group of H.G. Bock at Heidelberg University, Germany
for allowing us to use the software MUSCOD-II.
%
%
\bibliographystyle{IEEEtran}
\bibliography{manuscript_MannsICRAExo}
\end{document}